\documentclass[10pt,twocolumn,letterpaper]{article}

\usepackage[pagenumbers]{cvpr}

\usepackage{graphicx}
\usepackage{amsmath}
\usepackage{amssymb}
\usepackage{booktabs}
\usepackage{siunitx}
\usepackage{csquotes}
\usepackage{tabularx}

\usepackage[pagebackref,breaklinks,colorlinks]{hyperref}

\usepackage[capitalize]{cleveref}
\crefname{section}{Sec.}{Secs.}
\Crefname{section}{Section}{Sections}
\Crefname{table}{Table}{Tables}
\crefname{table}{Tab.}{Tabs.}

\def\confName{CVPR}
\def\confYear{2023}

\begin{document}

\title{An In-Depth Analysis of Adversarial Discriminative Domain Adaptation for Digit Classification}

\author{Eugene Choi$^*$\\
Princeton University\\
{\tt\small ec0342@princeton.edu}
\and
Julian Rodriguez$^*$\\
Princeton University\\
{\tt\small juliandr@princeton.edu}
\and
Edmund Young$^*$\\
Princeton University\\
{\tt\small edmund666@princeton.edu}
}

\maketitle
\def\thefootnote{*}\footnotetext{Authors contributed equally to this work}

\begin{abstract}

Domain adaptation is an active area of research driven by the growing demand for robust machine learning models that perform well on real-world data. Adversarial learning for deep neural networks (DNNs) has emerged as a promising approach to improving generalization ability, particularly for image classification. In this paper, we implement a specific adversarial learning technique known as Adversarial Discriminative Domain Adaptation (ADDA) and replicate digit classification experiments from the original ADDA paper. We extend their findings by examining a broader range of domain shifts and provide a detailed analysis of in-domain classification accuracy post-ADDA. Our results demonstrate that ADDA significantly improves accuracy across certain domain shifts with minimal impact on in-domain performance. Furthermore, we provide qualitative analysis and propose potential explanations for ADDA's limitations in less successful domain shifts. Code is \href{https://github.com/eugenechoi2004/COS429_FINAL}{here}.
\end{abstract}

\section{Introduction}
\label{sec:intro}

Machine learning models often struggle to generalize to new, unseen data due to differences in data distributions---a challenge known as \emph{domain shift}. In domain shift, a model trained on the \emph{source domain} fails to perform well when applied to the \emph{target domain}. This issue is pervasive in real-world machine learning applications, making it essential to develop robust models that can effectively adapt across domains for successful deployment.

To address domain shift, many domain adaptation methods focus on minimizing the discrepancy between the source and target domains~\cite{long2015learning,tzeng2014domain}. Adversarial adaptation methods accomplish this by training a model with an adversarial objective so that a discriminator is unable to distinguish data from the source and target domains. Tzeng \etal propose an unsupervised adversarial adaptation method called Adversarial Discriminative Domain Adaptation (ADDA)~\cite{tzeng2017adversarial}. \enquote{ADDA first learns a
discriminative representation using the labels in the source
domain and then a separate encoding that maps the target
data to the same space using an asymmetric mapping learned
through a domain-adversarial loss}~\cite{tzeng2017adversarial}.

\cite{tzeng2017adversarial} evaluates ADDA's performance on digit classification for domain shifts across the MNIST~\cite{lecun1998mnist}, USPS~\cite{hull1994usps}, and SVHN~\cite{netzer2011svhn} datasets. However, they only report out-of-domain accuracies for three of the possible six domain shift combinations with limited analysis of model interpretability. To bridge this gap, we implement ADDA from scratch and replicate their digit classification experiment on all six domain shifts. Furthermore, we provide post-ADDA in-domain accuracy results along with qualitative analysis using confusion matrices and t-SNE plots to explore potential shortcomings in classification performance. We finally provide possible hypotheses on several patterns we observe in the model decision-making process.

\section{Related Works}
\label{sec:relworks}

Previous work on transfer learning has been very popular for domain adaptation~\cite{kouw2019transfer}. Li \etal analyze the effects of stochastic feature augmentation (SFA)~\cite{li2021feataug} on domain adaptation by perturbing feature representations with both data-independent and adaptive Gaussian noise. Ganin \etal propose the domain-adversarial neural network (DANN)~\cite{ganin2016dann}, which modifies the training loss function by adding a domain adaptation regularization term to maximize prediction accuracy while remaining agnostic to the input data domain. The adversarial autoencoder (AAE) proposed by Makhzani \etal uses generative adversarial networks (GANs)~\cite{goodfellow2014gans} to train an encoder that \enquote{convert[s] the data distribution to the
prior distribution} and a decoder that \enquote{learns a deep generative model that maps the imposed prior to the data distribution}~\cite{makhzani2016aae}. Tobin \etal introduce domain randomization~\cite{tobin2017domainrand}, which simulates image data in different environments through randomized rendering. The goal of this method is to include enough variability so that real-world images are perceived as just another environment.

\section{Adversarial Discriminative Domain Adaptation (ADDA)}
\label{sec:adda}

Tzeng \etal generalize several state-of-the-art adversarial domain adaptation techniques including DANNs under a unified framework and propose ADDA~\cite{tzeng2017adversarial}, which we closely follow in our paper. ADDA is an unsupervised domain adaptation approach that uses a discriminative base model, unshared weights between the source and target mappings, and the standard GAN loss~\cite{tzeng2017adversarial} to learn an asymmetric mapping that matches the target to the source domain.

ADDA can be broken down into three steps: pre-training, adversarial adaptation, and testing. In the pre-training step, we first train a source encoder CNN and classifier on the source dataset with labels. Weights of both the source encoder and classifier are then frozen during the rest of the ADDA process. Next, we perform adversarial adaptation, where a target encoder is trained so that a discriminator is unable to tell which domain the input data is from. The purpose of this step is to train the target encoder to map input data to the shared feature space between the two domains. Note that both the discriminator and target encoder will be updated at each iteration during training in order for both components to adapt to each other. The final testing phase involves passing target domain data to the target encoder and original source classifier. The source classifier is only trained on data from the source domain and is agnostic to the target domain. For additional details on ADDA, refer to~\cite{tzeng2017adversarial}.

\section{Methodology}
\label{sec:methods}
\subsection{ADDA Implementation}

We implement ADDA as outlined in~\cite{tzeng2017adversarial} with PyTorch and use the same neural network architectures chosen for digit classification, consisting of a modified LeNet~\cite{lecun1998mnist,jia2014caffe} as the source and target encoders, a classifier with width $500$, and $3$ fully connected layers as the adversarial discriminator (the two hidden layers each have width $500$). The discriminator was initially trained with each hidden layer having only $100$ nodes to reduce training time, but due to poor performance, the original discriminator architecture was used instead. We implement the same standard GAN loss described in~\cite{tzeng2017adversarial} as the adversarial loss.

\subsection{Digits Datasets}

\begin{figure}[!h]
    \begin{subfigure}[!h]{0.2\textwidth}
        \includegraphics[width=\textwidth]{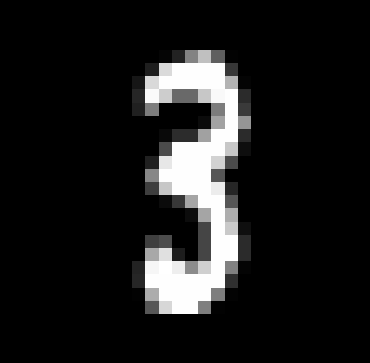}
        \caption{MNIST}
    \end{subfigure}
    \begin{subfigure}[!h]{0.2\textwidth}
        \includegraphics[width=\textwidth]{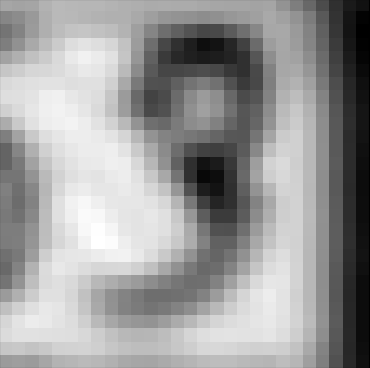}
        \caption{SVHN}
    \end{subfigure}
    \centering
    \begin{subfigure}[!h]{0.2\textwidth}
        \includegraphics[width=\textwidth]{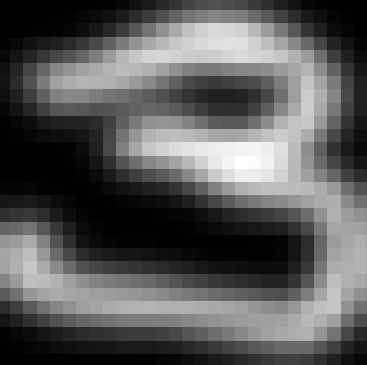}
        \caption{USPS}
    \end{subfigure}
    \caption{Images from the MNIST, SVHN, and USPS datasets of the digit $3$}
    \label{fig:mnist-svhn-usps}
\end{figure}

We use the three digits datasets used for digit classification in~\cite{tzeng2017adversarial} to analyze domain adaptation: MNIST~\cite{lecun1998mnist}, SVHN~\cite{netzer2011svhn}, and USPS~\cite{hull1994usps}. MNIST and USPS contain grayscale images of handwritten digits while SVHN contains real-world Google Street View images. Image samples from all three datasets can be seen in \cref{fig:mnist-svhn-usps}. We rescale images from all three datasets to $28\times28$ for consistency. We also convert SVHN data from RGB to grayscale for consistency with the other two datasets.

\subsection{Training}

We use mini-batch gradient descent with $200$ images per batch during training for both the source encoder/classifier and adversarial adaptation. While pre-training the source encoder, we use Adam~\cite{kingma2017adam} with a learning rate of $\num{1e-3}$. During adversarial adaptation training, we use Adam~\cite{kingma2017adam} with a learning rate of $\num{1e-4}$ and alternate training of the discriminator and target encoder every mini-batch iteration. For all six source-target domain shift combinations formed from the three datasets, adversarial adaptation training is done for $150$ epochs while source encoder training is done for $80$ epochs due to time limitations and limited computing resources. Note that this may affect classification performance as each source-target combination likely requires a different number of training epochs for optimal results. Hereinafter, we will refer to the trained source encoder and source classifier as the \emph{base model} and the trained target encoder and source classifier as the \emph{ADDA model}.

\section{Results}
\label{sec:results}
\subsection{Quantitative Analysis}

\begin{table}[!h]
    \centering
    \begin{tabularx}{0.4\textwidth}
    { | >{\centering\arraybackslash}X 
      | >{\centering\arraybackslash}X | }
    \hline
    \multicolumn{2}{|c|}{Preliminary Accuracies} \\
    \hline
    \textbf{Dataset} & \textbf{Accuracy} \\
    \hline   
    MNIST & $0.9879$ \\
    \hline
    SVHN & $0.8686$ \\
    \hline
    USPS & $0.9517$ \\
    \hline
    \end{tabularx}
    \caption{Preliminary accuracies for baseline models when evaluated in-domain}
    \label{tab:prelim_acc}
\end{table}

\begin{table}[h!]
    \centering
    \begin{tabularx}{0.4\textwidth}
    { | >{\centering\arraybackslash}X 
      | >{\centering\arraybackslash}X 
      | >{\centering\arraybackslash}X
      | >{\centering\arraybackslash}X | }
    \hline
    \multicolumn{4}{|c|}{Accuracies Across Domain Shifts} \\
    \hline
    \textbf{Source-Target} & \textbf{Baseline} & \textbf{ADDA-Target} & \textbf{ADDA-Source} \\
    \hline   
    MNIST-SVHN & $0.2374$ & $0.3292$ & $0.9784$ \\
    \hline
    MNIST-USPS & $0.4305$ & $0.6886$ & $0.9838$ \\
    \hline
    SVHN-MNIST & $0.5707$ & $0.6910$ & $0.3661$ \\
    \hline
    SVHN-USPS & $0.5969$ & $0.6059$ & $0.7775$ \\
    \hline
    USPS-MNIST & $0.5732$ & $0.7658$ & $0.9427$ \\
    \hline
    USPS-SVHN & $0.2567$ & $0.2542$ & $0.9412$ \\
    \hline
    \end{tabularx}
    \caption{Accuracies for all possible source-target combinations. The baseline column refers to the accuracy of the base model evaluated on test data from the target domain (without adversarial adaptation). The ADDA-target column refers to the accuracy of the ADDA model evaluated on test data from the target domain. The ADDA-source column refers to the accuracy of the ADDA model evaluated on test data from the source domain.}
\label{tab:accs}
\end{table}

\cref{tab:prelim_acc} provides the preliminary accuracies for each dataset when the base model is evaluated on test data from the source domain. Note that the base models trained on MNIST and USPS achieve significantly higher accuracies compared to SVHN. This is likely due to the fact that SVHN contains images where multiple digits may appear in the same image whereas MNIST and USPS have only one digit per image.

\cref{tab:accs} displays three different accuracies for each possible combination of source-target domains.

Comparing the baseline and ADDA-target accuracies, we find that ADDA improves generalization across all domain shifts except for USPS $\rightarrow$ SVHN. The largest improvement occurs in MNIST $\rightarrow$ USPS with a $0.2581$ increase in accuracy while the worst improvement occurs in USPS $\rightarrow$ SVHN for a $0.0025$ \emph{decrease} in accuracy. The two largest improvements in accuracy occur in MNIST $\rightarrow$ USPS and USPS $\rightarrow$ MNIST, which is expected as these two datasets are the most similar to one another. Interestingly, all domain shifts that contain MNIST as either the source or target show significant improvements compared to other source-target combinations. The average increase in accuracy among source-target combinations containing MNIST is $0.1657$, whereas the remaining two combinations have changes in accuracy of $0.009$ and $-0.0025$.

Looking at the ADDA-source column in \cref{tab:accs} and preliminary accuracies in \cref{tab:prelim_acc}, we find that after ADDA, in-domain accuracy decreases in all scenarios. This is due to the fact that ADDA maps input data to the shared feature space of both domains. However, among MNIST and USPS, this drop in accuracy is minimal (the largest decrease among the four domain shifts with either MNIST or USPS as the source is $0.0105$), while SVHN suffers large drops in accuracy (decreases of $0.5025$ and $0.0911$). This phenomenon is likely explained by the complexity of the SVHN dataset compared to MNIST and USPS. The SVHN feature space is probably very different compared to the feature spaces of both USPS and MNIST, and thus the target encoder is unable to map SVHN features onto the shared features space accurately for the classifier. Reasonably, we conclude that drastic domain shifts result in lower in-domain accuracy after ADDA training.

\subsection{Qualitative Analysis}

We visualize ADDA performance during test time for all combinations of source-target domain shifts except MNIST $\rightarrow$ USPS and USPS $\rightarrow$ MNIST (since the domains are quite similar) with confusion matrices and t-SNE plots~\cite{maaten2008tsne}.

\subsubsection{Confusion Matrices}
\label{subsec:conf}

\begin{figure}[!h]
    \begin{subfigure}[!h]{0.47\textwidth}
        \includegraphics[width=\textwidth]{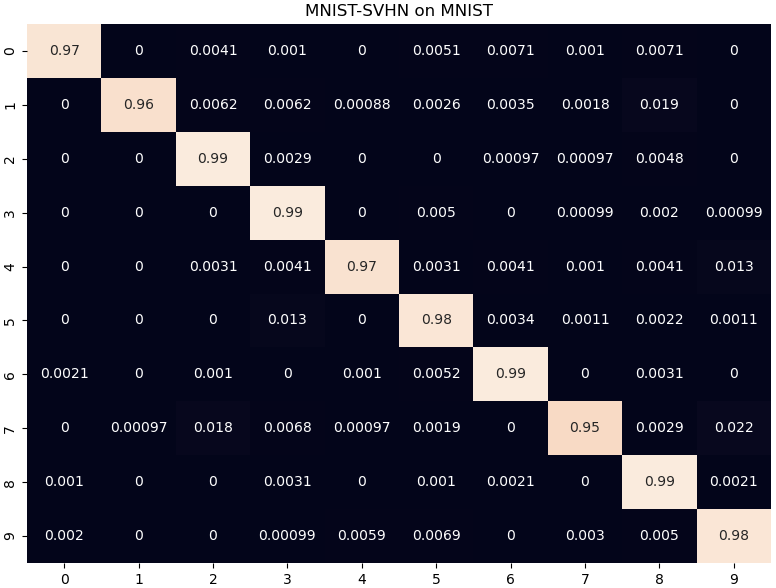}
        \caption{MNIST $\rightarrow$ SVHN evaluated on MNIST}
    \end{subfigure}
    \begin{subfigure}[!h]{0.47\textwidth}
        \includegraphics[width=\textwidth]{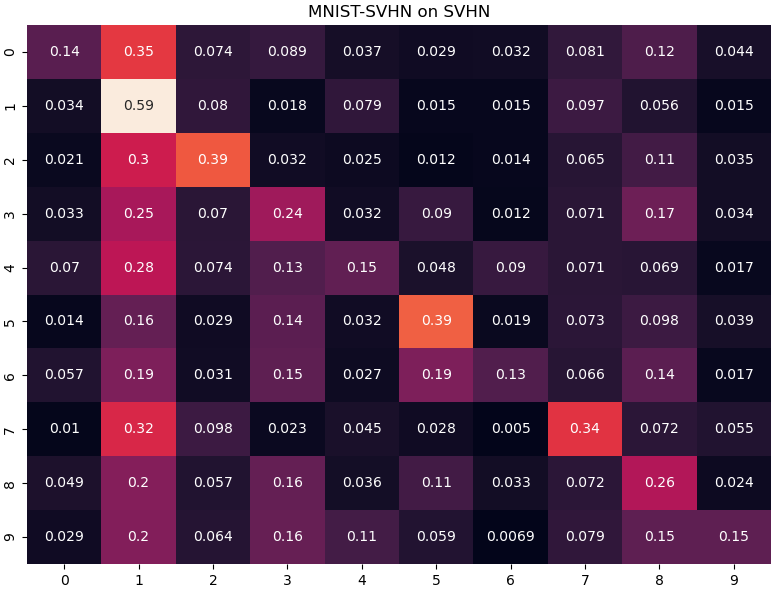}
        \caption{MNIST $\rightarrow$ SVHN evaluated on SVHN}
    \end{subfigure}
    \caption{MNIST $\rightarrow$ SVHN domain shift confusion matrices}
    \label{fig:conf-mnist-svhn}
\end{figure}

\begin{figure}[!h]
    \begin{subfigure}[!h]{0.47\textwidth}
        \includegraphics[width=\textwidth]{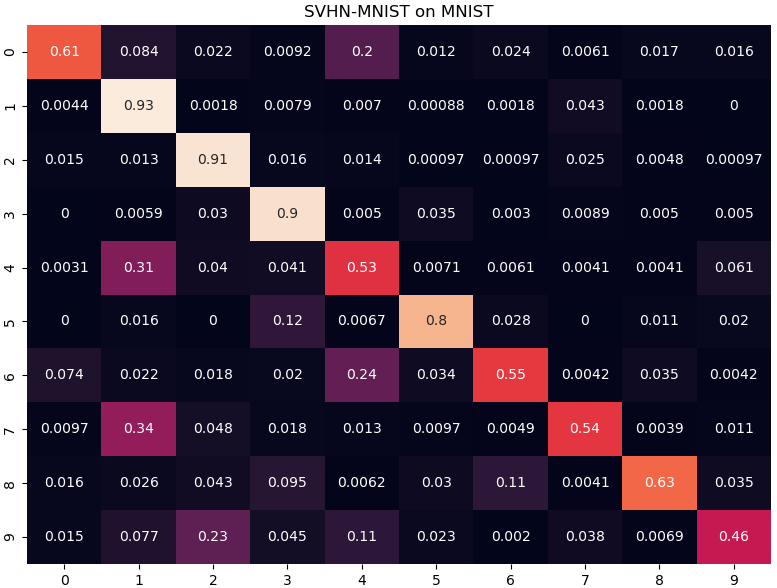}
        \caption{SVHN $\rightarrow$ MNIST evaluated on MNIST}
    \end{subfigure}
    \begin{subfigure}[!h]{0.47\textwidth}
        \includegraphics[width=\textwidth]{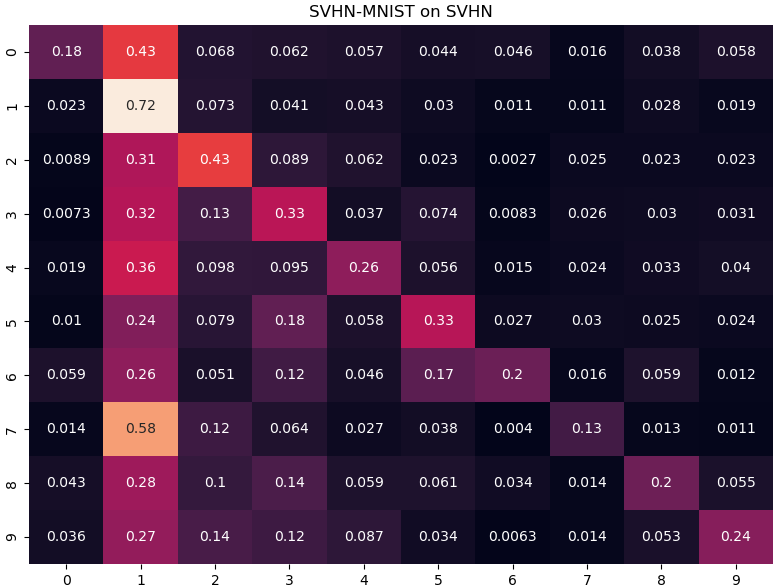}
        \caption{SVHN $\rightarrow$ MNIST evaluated on SVHN}
    \end{subfigure}
    \caption{SVHN $\rightarrow$ MNIST domain shift confusion matrices}
    \label{fig:conf-svhn-mnist}
\end{figure}

\begin{figure}[!h]
    \begin{subfigure}[!h]{0.47\textwidth}
        \includegraphics[width=\textwidth]{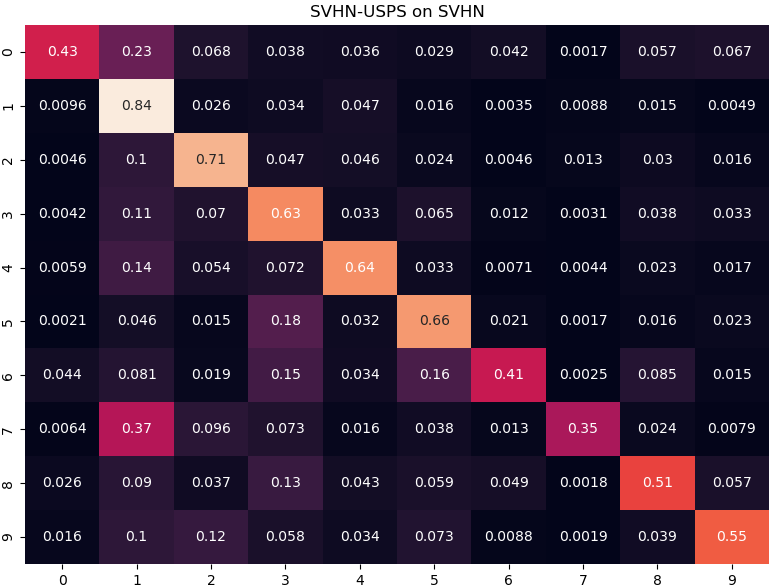}
        \caption{SVHN $\rightarrow$ USPS evaluated on SVHN}
    \end{subfigure}
    \begin{subfigure}[!h]{0.47\textwidth}
        \includegraphics[width=\textwidth]{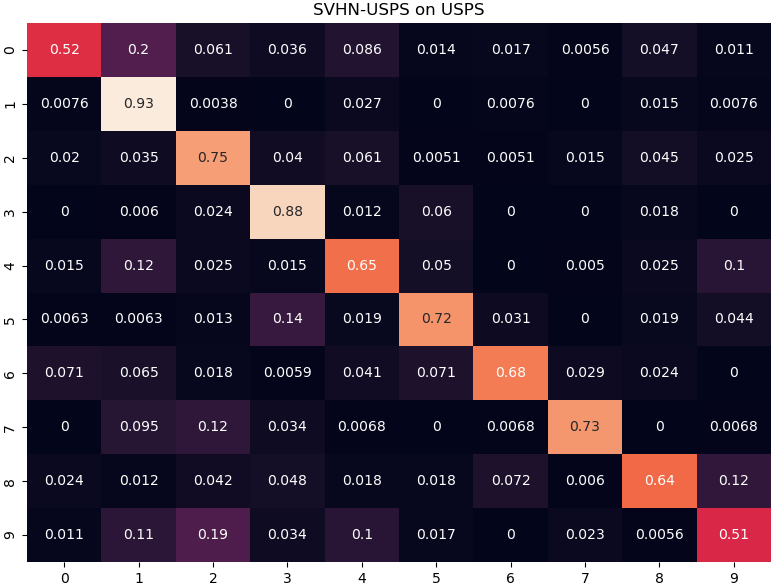}
        \caption{SVHN $\rightarrow$ USPS evaluated on USPS}
    \end{subfigure}
    \caption{SVHN $\rightarrow$ USPS domain shift confusion matrices}
    \label{fig:conf-svhn-usps}
\end{figure}

\begin{figure}[!h]
    \begin{subfigure}[!h]{0.47\textwidth}
        \includegraphics[width=\textwidth]{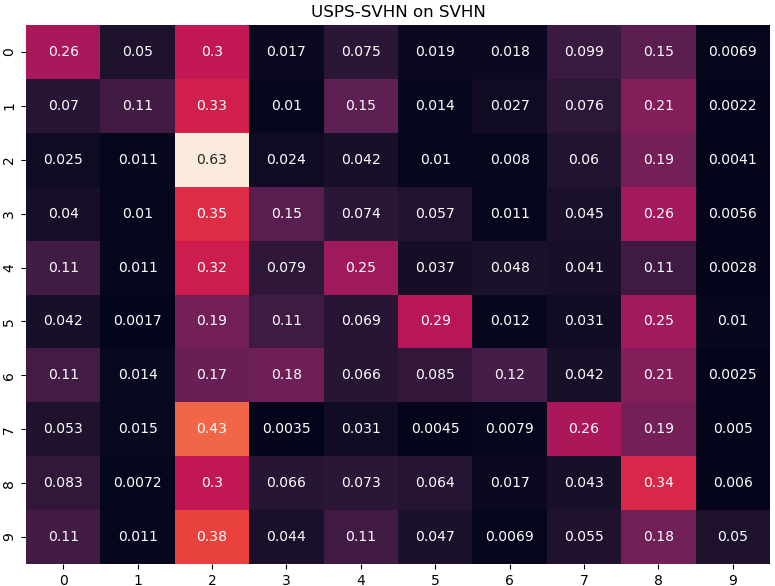}
        \caption{USPS $\rightarrow$ SVHN evaluated on SVHN}
    \end{subfigure}
    \begin{subfigure}[!h]{0.47\textwidth}
        \includegraphics[width=\textwidth]{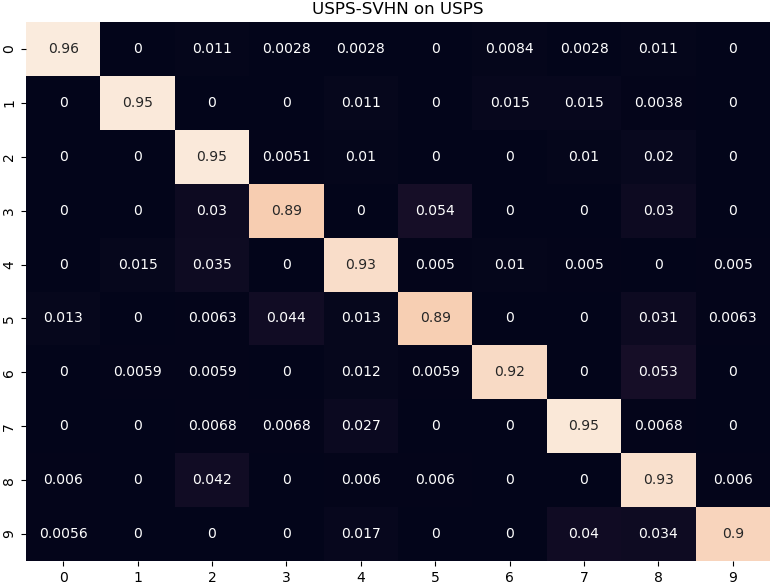}
        \caption{USPS $\rightarrow$ SVHN evaluated on USPS}
    \end{subfigure}
    \caption{USPS $\rightarrow$ SVHN domain shift confusion matrices}
    \label{fig:conf-usps-svhn}
\end{figure}

Analyzing \cref{fig:conf-mnist-svhn,fig:conf-svhn-mnist,fig:conf-svhn-usps,fig:conf-usps-svhn}, we find that in several of the confusion matrices, $7$ is often misclassified as $1$. This is likely due to the similar structure of both digits. We also notice that $0$ and $4$ are often misclassified as $1$. In general, $1$ is the most accurately predicted digit; this is probably due to the fact that $1$ is often predicted regardless of what the true label is. As a result, the classifier accurately predicts $1$'s at a higher frequency than other digits. Another possible explanation is that the digit $1$ has a relatively simple structure, and thus is easier to recognize.

Among the confusion matrices tested on SVHN, SVHN $\rightarrow$ USPS is the only domain shift whose classification output is not overrepresented by a certain digit. In the MNIST $\rightarrow$ SVHN and SVHN $\rightarrow$ MNIST domain shifts (lower panels in \cref{fig:conf-mnist-svhn,fig:conf-svhn-mnist}), $1$ is the most often predicted digit, while in the USPS $\rightarrow$ SVHN domain shift (upper panel in \cref{fig:conf-usps-svhn}), $2$ is the most often predicted digit.

Note that the USPS $\rightarrow$ SVHN domain shift evaluated on SVHN (upper panel in \cref{fig:conf-usps-svhn}) has consistently low accuracies (less than $0.4$) across all digits except for $2$, which has an accuracy of $0.63$ (this is probably due to $2$ being the most frequently predicted digit). This is also the only domain shift that suffers from a deterioration in out-of-domain accuracy after ADDA training (\cref{tab:accs}). One possible explanation for the poor performance is the constant number of training epochs enforced across all domain shifts; perhaps this particular domain shift requires more training time to adapt.

\subsubsection{t-SNE Plots}
\label{subsec:tsne}

\begin{figure}[!h]
    \begin{subfigure}[!h]{0.47\textwidth}
        \includegraphics[width=\textwidth]{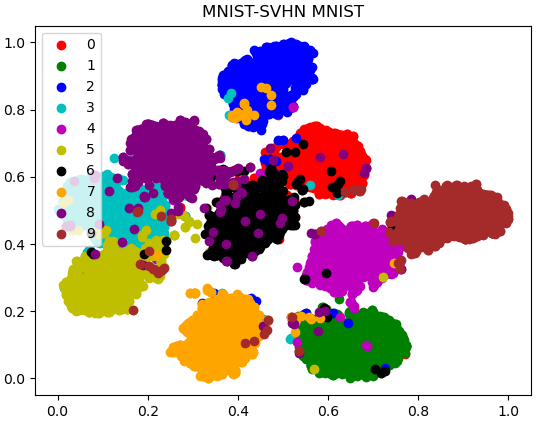}
        \caption{MNIST $\rightarrow$ SVHN evaluated on MNIST}
    \end{subfigure}
    \begin{subfigure}[!h]{0.47\textwidth}
        \includegraphics[width=\textwidth]{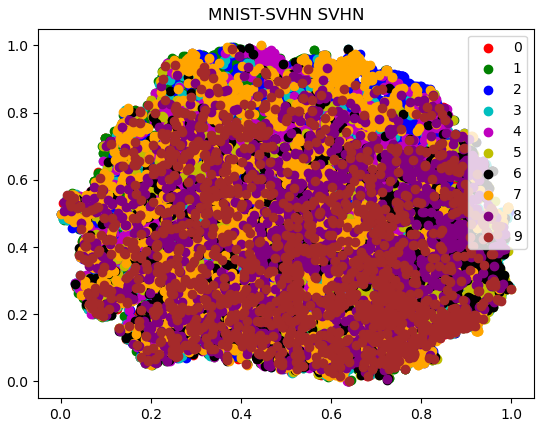}
        \caption{MNIST $\rightarrow$ SVHN evaluated on SVHN}
    \end{subfigure}
    \caption{MNIST $\rightarrow$ SVHN t-SNE Plots}
    \label{fig:tsne-mnist-svhn}
\end{figure}

\begin{figure}[!h]
    \begin{subfigure}[!h]{0.47\textwidth}
        \includegraphics[width=\textwidth]{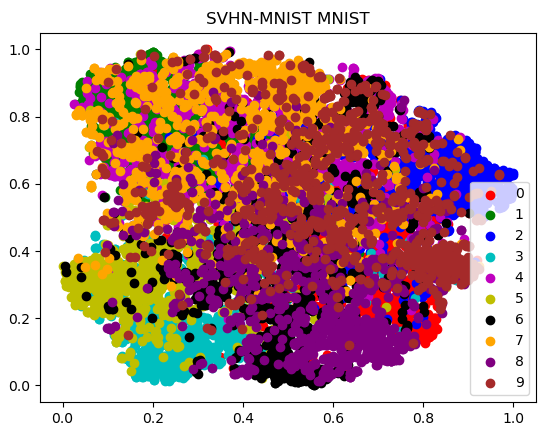}
        \caption{SVHN $\rightarrow$ MNIST evaluated on MNIST}
    \end{subfigure}
    \begin{subfigure}[!h]{0.47\textwidth}
        \includegraphics[width=\textwidth]{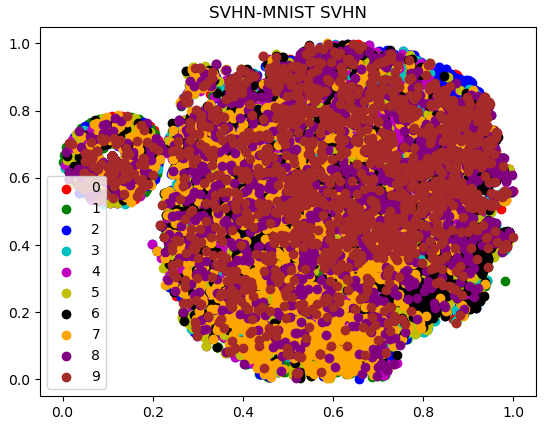}
        \caption{SVHN $\rightarrow$ MNIST evaluated on SVHN}
    \end{subfigure}
    \caption{SVHN $\rightarrow$ MNIST t-SNE Plots}
    \label{fig:tsne-svhn-mnist}
\end{figure}

\begin{figure}[!h]
    \begin{subfigure}[!h]{0.47\textwidth}
        \includegraphics[width=\textwidth]{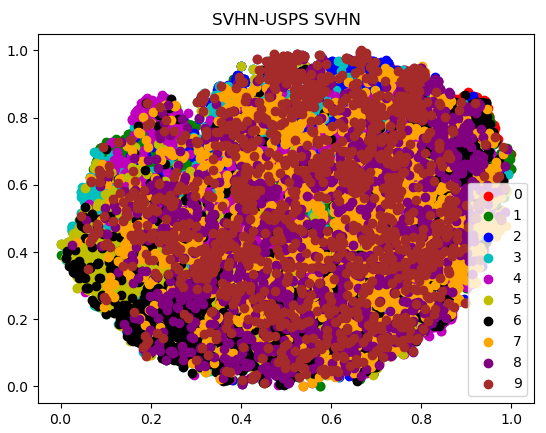}
        \caption{SVHN $\rightarrow$ USPS evaluated on SVHN}
    \end{subfigure}
    \begin{subfigure}[!h]{0.47\textwidth}
        \includegraphics[width=\textwidth]{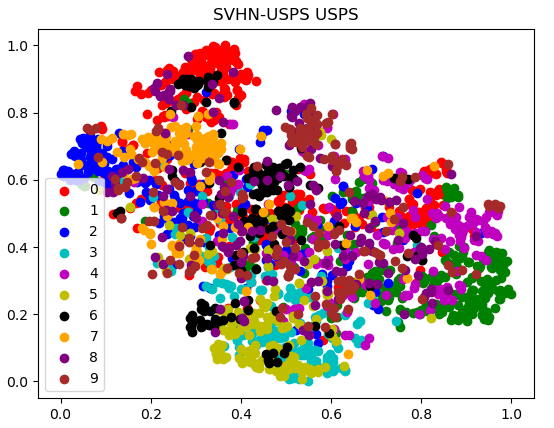}
        \caption{SVHN $\rightarrow$ USPS evaluated on USPS}
    \end{subfigure}
    \caption{SVHN $\rightarrow$ USPS t-SNE Plots}
    \label{fig:tsne-svhn-usps}
\end{figure}

\begin{figure}[!h]
    \begin{subfigure}[!h]{0.47\textwidth}
        \includegraphics[width=\textwidth]{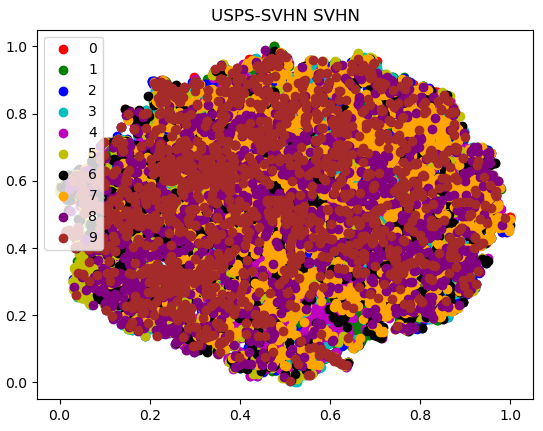}
        \caption{USPS $\rightarrow$ SVHN evaluated on SVHN}
    \end{subfigure}
    \begin{subfigure}[!h]{0.47\textwidth}
        \includegraphics[width=\textwidth]{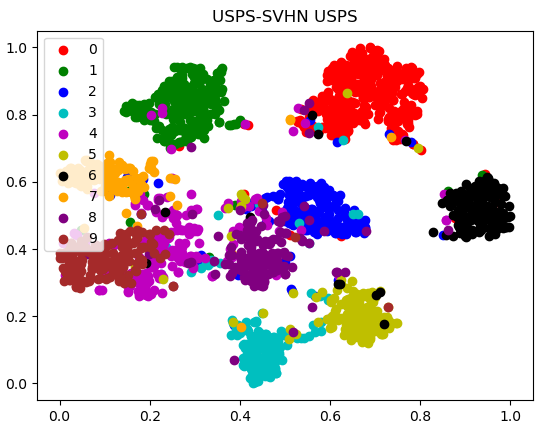}
        \caption{USPS $\rightarrow$ SVHN evaluated on USPS}
    \end{subfigure}
    \caption{USPS $\rightarrow$ SVHN t-SNE Plots}
    \label{fig:tsne-usps-svhn}
\end{figure}

\cref{fig:tsne-mnist-svhn,fig:tsne-svhn-mnist,fig:tsne-svhn-usps,fig:tsne-usps-svhn} display t-SNE plots~\cite{maaten2008tsne} of input data after they are passed through the target encoder but before the classifier.

As expected, we see clearly defined clusters for the MNIST $\rightarrow$ SVHN domain shift when evaluated on MNIST (upper panel in \cref{fig:tsne-mnist-svhn}) and USPS $\rightarrow$ SVHN when evaluated on USPS (lower panel in \cref{fig:tsne-usps-svhn}) since they have high classification accuracies. This indicates that the target encoder is able to learn meaningful representations and effectively differentiate digit features when data from the source domain is passed through. In both of these t-SNE plots, the cluster of $1$'s is close to the clusters of $4$'s and $7$'s, two digits that are often misclassified as $1$. $4$ and $9$ are also close to each other in both plots, as well as $3$ and $5$.

The SVHN $\rightarrow$ USPS domain shift evaluated on USPS (\cref{fig:tsne-svhn-usps} lower panel) shows rough clusters for most digits except for $4$ and $9$, which are instead scattered across the plot. In accordance with its confusion matrix (\cref{fig:conf-svhn-usps} lower panel), classification on the target domain is relatively successful compared to other domain shifts. Very rough clustering is also visible in the SVHN $\rightarrow$ MNIST domain shift evaluated on MNIST (\cref{fig:tsne-svhn-mnist} upper panel), with the digits $4$, $7$, $8$, and $9$ showing significant scattering. Despite the very rough clustering and scattering, we can see similar trends observed in the t-SNE plots with clear clusters: $3$ and $5$ are still close to each other, and $1$, $4$, and $7$ have significant overlap.

The remainder of the t-SNE plots do not show any distinguishable clustering, indicating that the target encoder is unable to extract identifying features of the different digits. Interestingly, the t-SNE plot of SVHN $\rightarrow$ MNIST evaluated on MNIST (\cref{fig:tsne-svhn-mnist} lower panel) contains a large central cluster and a much smaller cluster to the side, but both clusters appear to contain similar proportions of all digits.

We note that the t-SNE plots of domain shifts with SVHN as the target did not produce clusters only when evaluated on the SVHN dataset (\cref{fig:tsne-mnist-svhn,fig:tsne-usps-svhn}), indicating that adapting to SVHN as very difficult. This is probably due to the additional complexity of SVHN mentioned earlier.

\section{Future Work}
\label{sec:fut}

Computing power and time to train the models were the main limiting factors. With additional time and resources, we could have optimized model performance for each domain shift, allowing for more meaningful comparisons. This would have been most pertinent to the domain shifts that included SVHN which likely requires longer training times.

In addition, further qualitative analysis such as Grad-CAM~\cite{selvaraju2019gradcam} would be very insightful in providing visual explanations for poor model performance (particularly domain shifts evaluated on SVHN as the target). These visual explanations would be useful in understanding the current limitations of ADDA and potentially outline improvements to the training process.

ADDA can also be combined with domain randomization~\cite{tobin2017domainrand} to improve generalization across domain shifts. Instead of training the source encoder only on the source domain, one can also train it on simulated randomized data. Another approach is to train the target encoder on simulated data in addition to real data from the target domain.

We anticipate that the development of domain adaptation will be crucial for a wide variety of computer vision and robotics tasks. One application is Sim2Real research, which aims to use simulated instead of real-world data to train models due to data scarcity. Overcoming this domain shift could lead to promising results in reinforcement learning. Furthermore, demographic bias in facial recognition is a significant challenge where domain adaptation could be useful. Research shows facial recognition algorithms are more likely to misclassify faces with dark skin tones, in addition to women and minorities.

\section{Conclusion}
\label{sec:conclusion}

We have implemented and trained ADDA to successfully demonstrate its potential to improve digit classification accuracy across multiple domain shifts. In five of the six possible source-target combinations between the MNIST, USPS, and SVHN datasets, we find ADDA improves out-of-domain generalization ability, with significant improvements in four domain shifts. We also evaluate the ADDA-trained model in-domain and provide results on performance degradation compared to pre-ADDA in-domain accuracies. We find that there are minimal performance drops for MNIST and USPS, while SVHN suffers large performance drops.

Furthermore, we provide detailed qualitative analysis through confusion matrices and t-SNE plots. We then provide future possible avenues of research for domain adaptation.

\section*{Acknowledgments}

We would like to thank Prof. Olga Russakovsky for her advice and guidance. We would also like to acknowledge Princeton Research Computing for giving us access to the Adroit HPC cluster.

{\small
\bibliographystyle{unsrt}
\bibliography{references}

\begin{thebibliography}{10}

\bibitem{long2015learning}
Mingsheng Long, Yue Cao, Jianmin Wang, and Michael~I. Jordan.
\newblock Learning transferable features with deep adaptation networks, 2015.

\bibitem{tzeng2014domain}
Eric Tzeng, Judy Hoffman, Ning Zhang, Kate Saenko, and Trevor Darrell.
\newblock Deep domain confusion: Maximizing for domain invariance, 2014.

\bibitem{tzeng2017adversarial}
Eric Tzeng, Judy Hoffman, Kate Saenko, and Trevor Darrell.
\newblock Adversarial discriminative domain adaptation, 2017.

\bibitem{lecun1998mnist}
Y.~Lecun, L.~Bottou, Y.~Bengio, and P.~Haffner.
\newblock Gradient-based learning applied to document recognition.
\newblock {\em Proceedings of the IEEE}, 86(11):2278--2324, 1998.

\bibitem{hull1994usps}
J.J. Hull.
\newblock A database for handwritten text recognition research.
\newblock {\em IEEE Transactions on Pattern Analysis and Machine Intelligence}, 16(5):550--554, 1994.

\bibitem{netzer2011svhn}
Yuval Netzer, Tao Wang, Adam Coates, Alessandro Bissacco, Bo~Wu, and Andrew~Y. Ng.
\newblock Reading digits in natural images with unsupervised feature learning.
\newblock In {\em NIPS Workshop on Deep Learning and Unsupervised Feature Learning 2011}, 2011.

\bibitem{kouw2019transfer}
Wouter~M. Kouw and Marco Loog.
\newblock An introduction to domain adaptation and transfer learning, 2019.

\bibitem{li2021feataug}
Pan Li, Da~Li, Wei Li, Shaogang Gong, Yanwei Fu, and Timothy~M. Hospedales.
\newblock A simple feature augmentation for domain generalization.
\newblock In {\em 2021 IEEE/CVF International Conference on Computer Vision (ICCV)}, pages 8866--8875, 2021.

\bibitem{ganin2016dann}
Yaroslav Ganin, Evgeniya Ustinova, Hana Ajakan, Pascal Germain, Hugo Larochelle, François Laviolette, Mario Marchand, and Victor Lempitsky.
\newblock Domain-adversarial training of neural networks, 2016.

\bibitem{goodfellow2014gans}
Ian~J. Goodfellow, Jean Pouget-Abadie, Mehdi Mirza, Bing Xu, David Warde-Farley, Sherjil Ozair, Aaron Courville, and Yoshua Bengio.
\newblock Generative adversarial networks, 2014.

\bibitem{makhzani2016aae}
Alireza Makhzani, Jonathon Shlens, Navdeep Jaitly, Ian Goodfellow, and Brendan Frey.
\newblock Adversarial autoencoders, 2016.

\bibitem{tobin2017domainrand}
Josh Tobin, Rachel Fong, Alex Ray, Jonas Schneider, Wojciech Zaremba, and Pieter Abbeel.
\newblock Domain randomization for transferring deep neural networks from simulation to the real world, 2017.

\bibitem{jia2014caffe}
Yangqing Jia, Evan Shelhamer, Jeff Donahue, Sergey Karayev, Jonathan Long, Ross Girshick, Sergio Guadarrama, and Trevor Darrell.
\newblock Caffe: Convolutional architecture for fast feature embedding, 2014.

\bibitem{kingma2017adam}
Diederik~P. Kingma and Jimmy Ba.
\newblock Adam: A method for stochastic optimization, 2017.

\bibitem{maaten2008tsne}
Laurens van~der Maaten and Geoffrey Hinton.
\newblock Visualizing data using t-sne.
\newblock {\em Journal of Machine Learning Research}, 9(86):2579--2605, 2008.

\bibitem{selvaraju2019gradcam}
Ramprasaath~R. Selvaraju, Michael Cogswell, Abhishek Das, Ramakrishna Vedantam, Devi Parikh, and Dhruv Batra.
\newblock Grad-cam: Visual explanations from deep networks via gradient-based localization.
\newblock {\em International Journal of Computer Vision}, 128(2):336–359, October 2019.

\end{thebibliography}
}

\appendix

\end{document}